%% file: Main.tex
\documentclass[12pt, twocolumn]{NobArticle}
\footertext{\textit{Wingate University} (2025)}

\title{
    \begin{center}
        A Comparative Analysis of Principal Component Analysis (PCA) and Singular Value Decomposition (SVD) as Dimensionality Reduction Techniques
    \end{center}
}

\author{
    \begin{center}
        Michael Gyimadu \\
        Gregory Bell, Ph.D
    \end{center}
}

\date{
    \begin{center}
    Department of Mathematics, Wingate University \
    \end{center}
}

\renewcommand{\maketitlehookd}{%
\begin{abstract}
    \noindent High‑dimensional image data often require dimensionality reduction before further analysis. This paper provides a purely analytical comparison of two linear techniques—Principal Component Analysis (PCA) and Singular Value Decomposition (SVD). After the derivation of each algorithm from first principles, we assess their interpretability, numerical stability, and suitability for differing matrix shapes. We synthesize rule‑of‑thumb guidelines for choosing one out of the two algorithms without empirical benchmarking, building on classical and recent numerical literature. Limitations and directions for future experimental work are outlined at the end.
    
    \medskip

    \small{\textbf{Index Terms:} Principal Component Analysis, Singular Value Decomposition, Dimensionality Reduction, Eigendecomposition, Eigenvalues, Eigenvectors}
\end{abstract}
}

\begin{document}

\small
\maketitle

\input{Sections/01-Introduction}

\input{Sections/02-PCA}
\input{Sections/03-SVD}

\input{Sections/04-ComparativeAnalysis}
\input{Sections/05-Limitations}
\input{Sections/06-Conclusion}

\section*{Acknowledgements}
This project received support during the Linear Algebra course, instructed by Professor Gregory Bell, PhD at Wingate University, Spring 2025.

\end{document}

%% file: Sections/01-Introduction.tex
\section{Introduction}
High-dimensional datasets hinder statistical modeling, inflate storage costs, and makes it difficult to extract meaningful information from data. Dimensionality reduction addresses these issues by translating high-dimensional data (i.e thousands of pixel values in an image) into a more compact lower-dimensional feature space that still captures all the essential features of the image. Thus, instead of working with every raw dimension in the original dataset, we replace the ful-sized matrix/vector with a compressed one, while preserving structure. This has several benefits: smaller vectors/matrices mean faster algorithms and lower memory footprints; low variance/energy components are discarded which improves classification accuracy; and makes it easier to plot clusters and trends from 2- or 3-dimensional datasets. 

This study makes two contributions. First, we examine the linear‑algebra foundations of PCA and SVD highlighting their common structure and key differences. Second, we condense these theoretical insights into a brief section that addresses numerical robustness, aspect ratio, and data centering. No experiments are reported; instead, this is an analytical reference that will guide a future empirical benchmark on large-scale computer-vision datasets.

\subsection{Related Work}
Pearson's original formulation of PCA [1] and Eckart–Young’s Optimality Proof for the Truncated SVD [2] laid the groundwork for modern dimensionality reduction. Jolliffe and Cadima [4] survey statistical developments of PCA, while Golub and Reinsch [3] formalise numerically stable SVD implementations. This study synthesises insights from these strands but, unlike empirical surveys, remains fully theoretical.

\subsection{Image Representation as Matrices}
Matrix representation helps us to encode multiple features of an image into a matrix that mathematical operations can be performed on. For digital image representations, each element in the matrix corresponds to a pixel's color or intensity value.

Grayscale images can be represented as 2-dimensional matrices. The value of each element in the matrix corresponds to the intensity of the corresponding pixel, ranging from 0 (black) to 255 (white). Below is a 3x3 matrix representation of a grayscale matrix.
\[
    \begin{bmatrix}
        100 & 50 & 25 \\
        200 & 0 & 240 \\
        0 & 47 & 120
    \end{bmatrix}
\]

Color images, on the other hand, are typically represented as a 3-dimensional matrix or tensor. The first 2 dimensions of the tensor correspond to the height and width of the image; the third dimension represents the color channels. Each pixel is a vector of 3 values, one for each color in the Red-Green-Blue spectrum.

%% file: Sections/02-PCA.tex
\section{Principal Component Analysis (PCA)}
Principal Component Analysis achieves dimensionality reduction by transforming the original data matrix into a new coordinate system where the axes are called principal components.

\subsection{How PCA Works}
PCA works by finding the directions (principal components) along which the data varies the most. These principal components are orthogonalto each other, and the first principal component captures the most variance in the data, the second captures the second most variance and so on.

In other words, PCA takes a cloud of high-dimensional data points and rotates the entire space so that the new axes line up with the directions in which the data vary the most. After the rotation, the first axis (1st principal component) captures as much variance as possible, the second captures the largest spread left over and is perpendicular to the first, and so on. The amount of variance each axis explains is shown by its eigenvalue, so you can instantly see which directions matter and mostly describes noise.

\subsection{Implementation}
Suppose a matrix $X \in \mathbb{R}^{m\times n}$

We begin PCA by mean-centering the data, i.e center the data so each feature has mean 0.
\[
    X_c = X - X_\mu , X_c \in \mathbb{R}^{m\times n}
\]
where $X_\mu$: the matrix containing the means of each feature in $X$. We then form the covariance matrix $\mathbb{C}$ and do an eigencomposition of the matric, $\mathbb{C}$
\[
    \mathbb{C} = \frac{X_c^{T}X_c}{n - 1}, \mathbb{C} \in \mathbb{R}^{m\times }
\]
\[
    \mathbb{C} = \mathbb{Q}\Lambda\mathbb{Q}^T
\]
where $\mathbb{Q} \in \mathbb{R}^{m\times n}$ is an eigenvector matrix of $\mathbb{C}$ and the columns of $\mathbb{Q} = [q_1, q_2...q_d]$ are orthonormal eigenvectors (principal directions) and $\Lambda \in \mathbb{R}^{m\times n}$ is a diagonal matrix containing the eigenvalues (sample variances along the principal directions).

The eigenvalues correspond to the magnitude of the principal components - the larger the eigenvalue, the more important the corresponding principal component.

To reduce the dimensionality of a feature matrix then, you would pick the preferred principal components, say $s$. Now the principal component matrix is: $\mathbb{Q}_s \in \mathbb{R}^{n\times s} $

To project the data from the original data matrix onto the principal components, multiply the data matrix with the principal component matrix

\[
    \mathbb{X}_c\mathbb{Q}_s \in \mathbb{R}^{n\times s}
\]

%% file: Sections/03-SVD.tex
\section{Singular Value Decomposition}
SVD provides another way to factorize a matrix, into singular values and singular vectors. Singular value decomposition can be performed on any real matrix.

\subsection{Mathematical Framework}
Unlike eigendecomposition which involves analyzing a matrix $\mathbb{A}$ to discover a matrix V of eigenvectors and a vector of eigenvalues $\lambda$ such that 
\[
    \mathbb{A} = \mathbb{V}diag(\lambda)\mathbb{V}^{-1}
\]

SVD allows us to express $\mathbb{A}$ as
\[
    \mathbb{A} = \mathbb{U}\mathbb{D}\mathbb{V}^T
\]
Suppose that $\mathbb{A}$ is $m\times n$; then $\mathbb{U}$ is defined as an $m\times m$ matrix, $\mathbb{D}$, $m\times n$, and $\mathbb{V}$ to be $n\times n$.

The matrices $\mathbb{U}$ and $\mathbb{V}$ are both defined to be orthogonal matrices. The matrix $\mathbb{D}$ is defined to be a diagonal matrix. The elements along the diagonal of $\mathbb{D}$ are known as singular values of matrix $\mathbb{A}$. The columns of $\mathbb{U}$ are known as the left-singular vectors. the columns of $\mathbb{V}$ are known as the right-singular vectors.

In other words, $\mathbb{U}$ describes the main vertical patterns of the data. $\mathbb{V}$ captures the horizontal patterns, and $\mathbb{D}$ contains a row of weights (called singular values) that tell exactly how important each paired pattern is. The first few singular values (arranged from the largest to smallest) usually account for nearly all of the data's energy; the rest rail of toward noise.

In essence, dimensionality reduction with SVD just involves keeping the top $k$ most important pattern-weight pairs and discarding the rest. For any real matrix $\mathbb{X} \in \mathbb{R}^{m\times n}$, there exist orthogonal matrices $\mathbb{U} \in \mathbb{R}^{m\times m}, \mathbb{V} \in \mathbb{R}^{n\times n}$, and a diagonal vector $\Lambda$ such that
\[
    \mathbb{X} = \mathbb{U} \Lambda \mathbb{V}^T
\]
\[
    r=rank(x)
\]
\textit{Left singular vectors} form an orthonormal basis for the column space.
\textit{Right singular vectors} form an orthonormal basis for the row space.
Each singular value measures the "energy" carried by the vector pair.

%% file: Sections/04-ComparativeAnalysis.tex
\section{Comparative Analysis}
The table below summarises the theoretical trade‑offs that emerge from the preceding derivations.
\normalsize

\begin{table}[!htpb]
    \caption{Qualitative: summarises the theoretical trade‑offs that emerge from the preceding derivations.}
    \label{tab:table-01}
    \centering
    \begin{tabularx}{\linewidth}{l X X}
        \toprule
        \textbf{Criterion}& \textbf{PCA} & \textbf{SVD} \\  [0.25ex] 
        \midrule
        \textit{Data Centering Needed} & Yes & No \\
        \textit{Best for shape} & Near-square & Highly rectangular\\
        \textit{Interpretability} & Eigenvector variance & Energy captured by singular values \\
        \textit{Numerical Stability} & Moderate & High \\
        \bottomrule
    \end{tabularx}
\end{table}

%% file: Sections/05-Limitations.tex
\section{Limitations \& Future Directions}

This study provides no experimental validation. Future studies will perform an experimental evaluations of the two methods, PCA and SVD and benchmark them on standard computer‑vision datasets (mnist, cifar‑10, imagenet‑subset) using metrics such as explained variance, reconstruction error, runtime, and memory bandwidth under both CPU and GPU execution.

%% file: Sections/06-Conclusion.tex
\section{Conclusion}

This investigation set out to clarify when and why Principal Component Analysis (PCA) and Singular Value Decomposition (SVD) should be chosen for linear dimensionality reduction in image-based applications. 

Through a unified linear‑algebraic treatment, we have shown that PCA and SVD share a common optimisation framework yet respond differently to data centering, aspect ratio, and numerical conditioning. The qualitative guidelines distilled here helps to make a choice of which algorithm to use when only theoretical information is available. Subsequent empirical work will quantify these guidelines and test their robustness at scale on real-world data.